\documentclass[10pt,twocolumn,letterpaper]{article}

\usepackage[pagenumbers]{wacv} % To force page numbers, e.g. for an arXiv version

\usepackage{graphicx}
\usepackage{amsmath}
\usepackage{amssymb}
\usepackage{booktabs}

\usepackage{amsmath,amssymb,amsfonts}
\usepackage{graphicx}
\usepackage{textcomp}
\usepackage{xcolor}
\usepackage{flushend}
\usepackage{booktabs}
\usepackage{microtype}
\usepackage{graphicx}
 
\usepackage{booktabs} % for professional tables
\usepackage{resizegather}
\usepackage{subcaption}
\usepackage{algorithm,algpseudocode}
\usepackage{multirow}
\usepackage[pagebackref,breaklinks,colorlinks]{hyperref}

\usepackage[capitalize]{cleveref}
\crefname{section}{Sec.}{Secs.}
\Crefname{section}{Section}{Sections}
\Crefname{table}{Table}{Tables}
\crefname{table}{Tab.}{Tabs.}

\begin{document}

%%%%%%%%% TITLE - PLEASE UPDATE
\title{Adaptive Parametric Prototype Learning for \\ Cross-Domain Few-Shot Classification}

\author{Marzi Heidari, Abdullah Alchihabi, Qing En, Yuhong Guo\\
Carleton University, Canada\\
{\tt\small \{marziheidari@cmail., abdullahalchihabi@cmail., qingen@cunet., yuhong.guo@\}carleton.ca}
}
\maketitle

%%%%%%%%% ABSTRACT
\begin{abstract}
   Cross-domain few-shot classification induces a much more challenging problem than its in-domain counterpart due to the existence of domain shifts between the training and test tasks. 
In this paper, we develop a novel Adaptive Parametric Prototype Learning (APPL) method under the meta-learning convention for cross-domain few-shot classification. 
Different from existing prototypical few-shot methods that use the averages of support instances to calculate the class prototypes, we propose to learn class prototypes from the concatenated features of the support set in a parametric fashion and meta-learn the 
model by enforcing prototype-based regularization on the query set. 
In addition, we fine-tune the model in the target domain in a transductive manner using a weighted-moving-average self-training approach on the query instances. 
We conduct experiments on multiple cross-domain few-shot benchmark datasets. 
The empirical results demonstrate that APPL yields superior performance than many state-of-the-art cross-domain few-shot learning methods.
\end{abstract}
\section{Introduction}
Benefiting from the development of deep neural networks, 
significant advancement has been achieved on image classification 
with large amounts of 
annotated data.
However, obtaining large amounts of annotated data is time-consuming and labour-intensive, 
while it is difficult to generalize trained models to new categories of data.
As a solution, few-shot learning (FSL) has been proposed to classify instances from unseen classes using only a few labeled instances.
FSL methods usually use a base dataset with labeled images to train 
a prediction model in the training phase.
The model is then fine-tuned on the prediction task of
novel categories with a few labeled instances (i.e. support set), and finally evaluated on the test data (i.e. query set) from the same novel categories in the testing phase.
FSL has been widely studied in the in-domain settings where 
the training and test tasks are from the same domain
\cite{finn2017model,snell2017prototypical,lee2019meta}.
However, when the training and test tasks are in different domains, 
it poses a much more challenging cross-domain few-shot learning problem than its in-domain counterpart 
due to the domain shift problem.

Recently, several methods have made progress to address cross-domain few-shot learning,
including the ones based on
data augmentation, data generation \cite{advTaskAug,yeh2020large,islam2021dynamic} 
and self-supervised learning \cite{phoo2020self} techniques. 
However, such data generation and augmentation methods increase the computational cost and 
cannot scale well to scenarios with higher-shots \cite{advTaskAug}. 
Some other works either require 
large amounts of labeled data from multiple source domains \cite{hu2021switch} 
or the availability of substantial unlabeled data from the target domain 
during the source training phase \cite{phoo2020self,islam2021dynamic,yao2021cross}.
Such requirements are hard to meet and hence hamper their applicability in many domains. 
Although some existing prototypical-based few-shot methods have also been applied to 
address cross-domain few-shot learning due to their simplicity and computational efficiency \cite{snell2017prototypical,garcia2018fewshot}, 
these standard methods lack sufficient capacity in handing large cross-domain shifts 
and adapting to target domains.

In this paper, we propose a novel Adaptive Parametric Prototype Learning (APPL) method 
under the meta-learning convention for cross-domain few-shot image classification.
APPL introduces a parametric prototype calculator network (PCN) to learn class prototypes 
from concatenated feature vectors of the support instances
by ensuring the inter-class discriminability and intra-class cohesion with prototype regularization losses. 
The PCN is 
meta-learned on the source domain using the labeled query instances. 
In the target domain, 
we deploy a weighted-moving-average (WMA) self-training approach to leverage the unlabeled query instances 
to fine-tune the prototype-based prediction model in a transductive manner.
With PCN and prototype regularizations, 
the proposed method is expected to have better generalization capacity in 
learning class prototypes in the feature embedding space,
and hence effectively mitigate the domain shift and adapt to the target domain with WMA self-training.  
Comprehensive experiments are conducted on eight cross-domain few-shot learning benchmark datasets.
The empirical results demonstrate the efficacy of the proposed APPL 
for cross-domain few-shot classification by comparing with 
existing state-of-the-art methods.

%%%%%%%%%%%
 \begin{figure*}[t]
  \centering
 \includegraphics[scale=0.55]{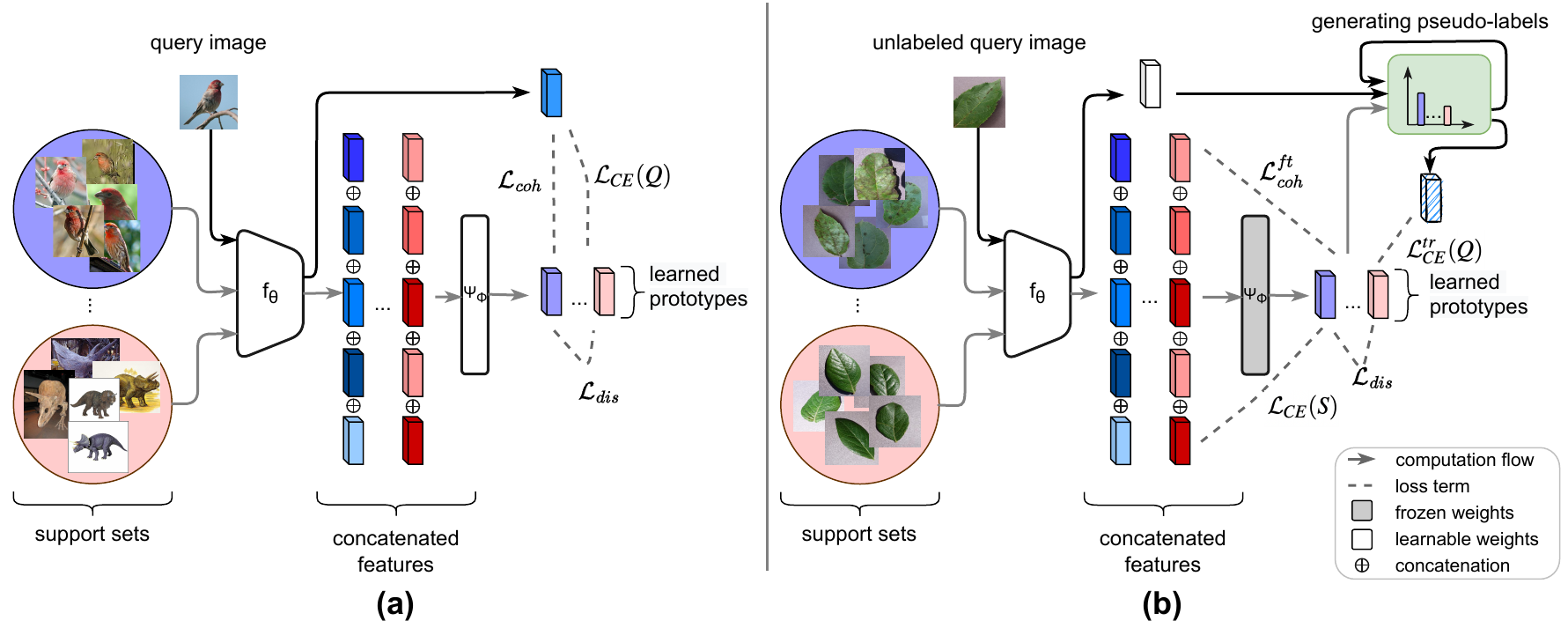} 
	 \caption{The proposed APPL method. \textbf{(a) Training on the source domain}.
	 The concatenated feature vectors of each class are fed into the PCN ($\psi_\phi$) 
	 to produce class prototypes, which are used to compute the meta-training loss terms on query instances. 
	 \textbf{(b) Fine-tuning on the target domain.}  
	 Both the labelled support set and the unlabeled query set with soft pseudo-labels computed
	 using WMA are used to fine-tune the feature encoder in the target domain with prototype based losses. 
	 }
   \label{fig:target}
\end{figure*}
%%%%%%%%%%%%%%%%%%%%%%%%%%%%%%%%%%
\section{Related Works}
\subsection{Few-Shot Learning}
Most FSL studies have focused on the in-domain settings. 
The FSL approaches 
can be grouped into
three main categories: metric-based and meta-learning approaches \cite{finn2017model,snell2017prototypical,lee2019meta}, transfer learning approaches \cite{guo2019spottune,jeong2020ood,ge2017borrowing,yosinski2014transferable,dhillon2019baseline} and augmentation and generative approaches \cite{zhang2018metagan,lim2019fast,hariharan2017low,schwartz2018delta,reed2017few}. 
In particular, the representative meta-learning approach, MAML \cite{finn2017model}, 
learns good initialization parameters from various source tasks that make the model easy to adapt to new tasks. 
The non-parametric metric-based approach, 
MatchingNet \cite{vinyals2016matching},
employs attention and memory in order to train a network that learns from few labeled samples. ProtoNet \cite{snell2017prototypical} learns a metric space where each class is represented by the average of 
the available support instances and classifies query instances based on their distances to the class prototypes. 
A few meta-learning works, such as RelationNet \cite{sung2018learning}, GNN \cite{garcia2018fewshot} and Transductive Propagation Network (TPN) \cite{liu2018learning}, 
exploit the similarities between support and query instances to classify the query instances.  
MetaOpt uses meta-learning to train a feature encoder that obtains discriminative features for a linear classifier \cite{lee2019meta}. 
Transfer learning methods initially train a model on base tasks 
and then use various fine-tuning methods to adapt the model to novel tasks 
\cite{guo2019spottune,jeong2020ood,ge2017borrowing,yosinski2014transferable,dhillon2019baseline}. 
Generative and augmentation approaches 
generate additional samples to increase the size of available data during training \cite{zhang2018metagan,lim2019fast,hariharan2017low,schwartz2018delta,reed2017few}.

\subsection{Cross-Domain Few-shot Learning}

Recently cross-domain few-shot learning (CDFSL) has started receiving more attention \cite{guo2020broader,phoo2020self}.
Tseng \etal \cite{Tseng2020CrossDomain} propose a feature-wise transformation (FWT) layer that is used jointly with standard few-shot learning methods for cross-domain few-shot learning. 
The FWT layer uses affine transformations to augment the learned features in order to help the trained network generalize across domains. 
Du \etal \cite{du2022hierarchical} address the domain shift problem by proposing a prototype-based Hierarchical Variational neural Memory framework (HVM), where the hierarchical prototypes and the memory are both learned using variational inference. 
Adler \etal \cite{adler2020cross} propose a Cross-domain Hebbian Ensemble Fusion (CHEF) method, which applies an ensemble of Hebbian learners on different layers of the neural network to obtain a representation fusion. 
Data augmentation and data generation methods have also been used to bridge the gap between the source and target domains \cite{advTaskAug,yeh2020large,islam2021dynamic}. 
Wang \& Deng \cite{advTaskAug} propose an Adversarial Task Augmentation approach (ATA) to generate difficult training tasks in an adversarial fashion and improve the generalizability of few-shot methods across largely different domains. 
Islam \etal \cite{islam2021dynamic} employ dynamic distillation and consistency regularization to train student and teacher networks jointly 
on the source domain data and unlabeled data from the target domain. 
Hu \etal \cite{hu2021switch} propose domain-switch learning framework with multiple source domains, and use re-weighted cross-entropy loss and binary KL divergence loss to prevent overfitting and catastrophic forgetting. 
Sun \etal \cite{sun2021explanation} adapt the explanation method of Layer-wise Relevance Propagation (LRP) to the FSL setup, which 
guides the FSL training 
by dynamically highlighting the discriminative features of the input samples.

Some other works \cite{triantafillou2019meta,liu2021multi,bateni2020improving,doersch2020crosstransformers} 
have tested alternative cross-domain few-shot learning settings such as Meta-Dataset \cite{triantafillou2019meta} CDFSL setting where models are trained on several source-domain datasets and tested on multiple target-domain datasets. 
In this work, we focus on the most widely studied CDFSL setting adopted in the work of \cite{guo2020broader}.

%%%%%%%%%%%%%%%%%%%%%%%%%%%%%%%%%%%%%%%%%%
\section{Approach}
\label{gen_inst}

\subsection{Preliminary}
The cross-domain few-shot learning problem aims to train a model on the source domain with its large set of labelled instances 
and then adapt the model to address the prediction task in the target domain with few labeled instances. 
We assume the two domains have different distributions in the input space ($\mathcal{P}_s \neq \mathcal{P}_t$) 
and have disjoint classes 
($\mathcal{Y}_s \cap \mathcal{Y}_t=\emptyset$).
In the target domain, the model is provided with a support set $S=\{(x_i,y_i)\}^{N_s}_{i=1}$ and tested on a query set $Q=\{(x_i,y_i)\}^{N_q}_{i=1}$ where $N_s$ and $N_q$ are the sizes of the support and query sets respectively. 
The support set is made up of $N$ classes with $K$ instances in each class, which is commonly described as N-way K-shot. 

In the classic prototypical few-shot learning \cite{snell2017prototypical},
each image $x$ first goes through a 
feature encoder $f_{\theta}$ and obtains its embedding vector in the feature space.
Then, for each class in the support set, 
a prototype $p_n \in \mathbb{R}^D$ is computed as 
the average embedding vector of the support instances: 
 $           p_n = \frac{1}{K} \sum\nolimits_{(x,y) \in S_n}f_{\theta}(x)$,
where 
$S_n$ denotes the set of $K$ support instances from class $n$. 
To classify the query instances, the distances between each query sample and the prototypes of all classes in the support set are computed.
Then the softmax function is used to normalize the calculated distances to obtain 
the class prediction probabilities as follows: 
\begin{equation}
   P(y=j|x)=\frac{\exp(-d(f_\theta(x),p_{j}))}{\sum_{n=1}^{N}\exp(-d(f_{\theta}(x),p_n))},
\end{equation}
where $d(.,.)$ is a distance function and $P(y=j|x)$ is the predicted probability that query sample $x$ 
belonging to class $j$. 
During the meta-training phase, the model is trained to minimize the cross-entropy loss 
on the query instances: 
\begin{equation}
	\mathcal{L}_{CE}(Q) =  \sum\nolimits_{x\in Q} \ell_{CE}(P_x,Y_x),
\end{equation}
where $\ell_{CE}$ is the cross-entropy function, $P_x$ and $Y_x$ are the predicted class probability vector 
and ground-truth label indicator vector respectively for a query sample $x$.

%%%%%%%%%%%%%%%%%%%%%%%%%%%%%%%%%%%%%%%
\subsection{Adaptive Parametric Prototype Learning}
In this section, we present our proposed Adaptive Parametric Prototype Learning (APPL) method
for cross-domain few-shot image classification. 
The overall framework of APPL is illustrated in Figure \ref{fig:target}. 
APPL first performs meta-training in the label-rich source domain by meta learning an adaptive prototype calculator network (PCN)
after the feature encoder.  
PCN generates the prototype of each class 
by segregating information
from the concatenated feature vectors of the K-shot support instances. 
Then the trained model can be fine-tuned for the few-shot classification task
in the target domain using a weighted-moving-average (WMA) self-training approach, 
which aims to adapt the model to the target domain and 
further improve the quality of the learned class prototypes.
We describe the details below.

%%%%%%%%%%%%%%%

%%%%%%%%%%%%%%%%%%%%%%%%%%%%%%%%%%%%%%%%%%%%%%%%%%%%%
\subsubsection{Adaptive Prototype Calculator Network}
Simply averaging the support instances to calculate the class prototypes has
the evident drawback of ignoring 
the inter-class and intra-class instance relations.
To overcome this drawback, we propose to learn class prototypes from 
the support instances through a parametric prototype calculator network
by enforcing both inter-class discriminability and intra-class cohesion in the extracted feature space.
Such a parametric prototype generation mechanism is expected to 
produce more representative class prototypes from various support instance layouts,
and guide the feature encoder to better adapt to the target domain through fine-tuning in the testing phase. 

We define the adaptive prototype calculator network (PCN) as $\psi_{\phi} : \mathbb{R}^{K \cdot D} \rightarrow \mathbb{R}^D$, where $D$ is the size of the learned embeddings by the feature encoder, $f_{\theta}$, 
$K$ is the number of support instances per class,
and $\phi$ denotes the parameters of the PCN. 
Specifically, PCN
takes the concatenated feature vectors of the support instances of a given class
as input, and outputs the prototype of the corresponding class: 
\begin{equation}\label{eq:prototypes}
    p_{n} = \psi_{\phi} (concat (f_{\theta}(x^n_1),..,f_{\theta}(x^n_K))  ),
\end{equation}
where $x^n_j$ denotes the $j$-th support instance from class $n$ and $p_{n}$ is the learned prototype of class $n$.
By feeding the support instances of each class to $\psi_{\phi}$, 
we can obtain the prototypes for all the $N$ classes: $\mathbb{P}=\{p_1,p_2,..,p_N\}$.

We train the PCN during the meta-training phase using the few-shot training tasks in the source domain. 
Specifically, given the feature encoder trained on the support instances, we update the 
parameters of the PCN 
by minimizing the cross-entropy loss on the query instances, $\mathcal{L}_{CE}(Q)$. 
Moreover, 
we introduce two auxiliary regularization loss terms, a prototype discriminative loss 
and a prototype cohesive loss, 
to ensure the learned prototypes are both discriminative and representative of the underlying classes.
To elaborate, the prototype discriminative loss 
$\mathcal{L}_{dis}$ 
aims to push the prototypes of different classes away from each other
and is defined as follows:
\begin{equation}
\label{eq:dis}
\mathcal{L}_{dis} =  \frac{1} {\sum_{\{p_i,p_j\} \in \mathbb{P}} \, \, d(p_i,p_j)},
\end{equation}
We in particular use a squared Euclidean distance as $d(\cdot,\cdot)$.
By contrast, the prototype cohesive loss $\mathcal{L}_{coh}$ is designed to 
pull the prototypes and the query instances of their corresponding classes to be closer to each other:
\begin{equation}
\label{eq:coh}
\mathcal{L}_{coh} = \sum\nolimits_{n=1}^{N}\sum\nolimits_{x \in Q_{n}} d(p_n,f_\theta(x) ),
\end{equation} 
where $Q_{n}$ 
denotes the set of query instances from class $n$.
Overall, PCN is meta-trained in the source domain by minimizing the following joint loss:
\begin{equation}\label{eq:train_loss}
     	\min_{ \phi}\;   
	\mathcal{L}_{train} 
	= \mathcal{L}_{CE}(Q) + \lambda_{dis} \mathcal{L}_{dis} +\lambda_{coh}  \mathcal{L}_{coh},
\end{equation}
where 
$\lambda_{dis}$ and $\lambda_{coh}$ are the trade-off hyper-parameters 
that control the contribution of 
the two regularization loss terms, 
$\mathcal{L}_{dis}$ and $\mathcal{L}_{coh}$, respectively. 
The meta-training procedure for the proposed
APPL is summarized in Algorithm \ref{alg:source}. 
\begin{algorithm}[tp]
  \caption{Training Procedure on Source Domain}
  \begin{algorithmic}
    \State \textbf{Input}: Source domain dataset $D_{s}$, $K$, $N$;\\ \quad \quad \quad pre-trained feature extractor $ f_\theta $; \\
    \quad \quad  \quad  
    learning-rate $\gamma_1$ and $\gamma_2$;\\
       \quad \quad  \quad  initialize: $\phi \leftarrow \phi_0,\theta \leftarrow \theta_0;$
    \State \textbf{Output}: Learned model parameters  $\theta, \phi$
  
     \For{{iter = 1} {\bf to} maxiters} 
        \State $V \leftarrow $ \text{randomly sample N class indices}\\ \quad  \quad  \quad \text{from all classes in} $D_{s} $ 
      \For{$n$ in $\{1,..,N\}$}
        \State $S_n,Q_n \leftarrow$ \text{randomly sample 
        support \& query sets}\\ 
        \quad  \quad  \quad \quad  \text{for class $n$ from } $ D_{s}^V$
  
      \EndFor

      \State $S=S_1 \cup .. \cup S_N$, \; $Q=Q_1 \cup .. \cup Q_N$    
	  \For{{initer=1} {\bf to} maxiniters}
           \State $ \theta   \leftarrow  \theta   - \gamma_1\nabla_{ \theta} \mathcal{L}_{CE}  (S)$
    	  \EndFor

	  \State Compute $\mathcal{L}_{dis}$ and $\mathcal{L}_{coh}$ with Eq. (\ref{eq:dis}) and Eq.(\ref{eq:coh}) 
	  \State $\mathcal{L}_{train}=\lambda_{dis}\mathcal{L}_{dis}+\lambda_{coh}\mathcal{L}_{dis}$ 
	  \For{$(x,y)\in Q$}
    \State$\mathcal{L}_{train} \leftarrow \mathcal{L}_{train} + \ell_{CE}(x,y)$
        \EndFor
   \State $ \phi   \leftarrow  \phi   - \gamma_2  \nabla_\phi\mathcal{L}_{train}$
  \EndFor
  \end{algorithmic}
\label{alg:source}

\end{algorithm}
  
%%%%%%%%%%%%%%%%%%%%%%%%%%%%%%%%%%%%%%%%%%%%%%%%%%%%% 
\subsubsection{Weighted Moving Average Self-Training}

For cross-domain few-shot image classification, 
significant distribution discrepancies in the input image space typically exist between the source and target domains. 
Hence after meta-training the feature encoder $f_\theta$ and the PCN $\psi_{\phi}$ in the source domain, 
it is critical to fine-tune 
the feature encoder $f_\theta$ on the few-shot test task in the target domain 
to overcome the cross-domain gap as well as adapt $f_\theta$ to the target test task. 
Due to the scarcity of the labeled support instances in the target task, we propose to 
employ the unlabeled query instances with predicted soft pseudo-labels to 
increase the size and diversity of the target data for fine-tuning and
mitigate the domain shift between the source and target domains. 
To this end, we develop 
a weighted-moving-average (WMA) self-training approach 
to compute the soft pseudo-labels and deploy the query instances for fine-tuning. 

Specifically, at each iteration $i$ of the fine-tuning process, 
we first calculate the distances between each query instance $x$ and the class prototypes 
$[p^i_1, p^i_2,\cdots, p^i_N]$
produced by the PCN $\psi_{\phi}$ from the support set $S$ for all the $N$ classes, 
and form the following distance vector for $x$: 
\begin{equation}
    h^{i}(x) = [d(f_{\theta^i}(x),p^{i}_{1}), d(f_{\theta^i}(x),p^{i}_{2}),..,d(f_{\theta^i}(x),p^{i}_{N})  ]^{\top}
    \label{eq:hx}
\end{equation}
Then we use this distance vector $h^{i}(x)$ 
to perform weighted moving average update 
and maintain a weighted-moving-average distance
vector $\tilde{h}^{i}(x)$ 
for the current iteration $i$ 
as follows:
\begin{equation}
   \tilde{h}^{i}(x) =   \alpha_i \; h^{i}(x)+ (1-\alpha_i) \;   \tilde{h}^{i-1}(x),
	\label{eq:tildehx}
\end{equation}
where $\alpha_i$ is a trade-off parameter that controls the combination weights between 
distances computed from the current iteration
and previous iterations.  
The weighted-moving-average distance vectors can then be used to 
compute the class prediction probabilities over each query instance $x$
by using the softmax function:
\begin{equation}\label{eq:pseudo}
  \tilde{P}^{i}(y=j|x)=\frac{\exp(-\tilde{h}^{i}
  _{j}(x))}{\sum_{n=1}^{N}\exp(-\tilde{h}_{n}^{i}(x))},
 \end{equation}
where $\tilde{P}^{i}(y=j|x)$ is the probability of the query instance $x$ 
being assigned to class $j$ at iteration $i$. 
By using the predicted class probabilities as soft pseudo-labels,
the query instances 
can subsequently be used to support 
the fine-tuning of $f_{\theta}$ in a self-training manner. 
The weighted-moving-average (WMA) update mechanism can stabilize the self-training process and dampen possible oscillating predictions for challenging query instances.
Moreover, to increase 
the stability and convergence property of the WMA self-training, 
we adopt the following rectified annealing schedule for the WMA hyper-parameter $\alpha_i$: 
\begin{equation}
    \alpha_i =  \gamma\, \alpha_{i-1},
	\label{eq:alpha}
\end{equation}
where 
$\gamma\in(0,1)$ is a reduction ratio 
parameter used for updating the $\alpha$ value across iterations. 
This annealing schedule can enable
larger updates to the $\tilde{h}^{i}$ vectors   
in the beginning iterations of fine-tuning by starting with a large value $\alpha_0$, 
while gradually reducing the degree of update with the decreasing of $\alpha_i$ 
in later iterations. 

%%%%%%%%%%%%%%%%%%%%%%%%%%%%%%%%%%%%%%%%%%%%%%%%%%
\begin{algorithm}[t]
  \caption{Fine-tuning Procedure on Target Domain}
  \begin{algorithmic} 
    \State \textbf{Input}: 
	  Target N-way-K-shot test task $(S, Q)$;\;\\ \quad  \quad  \quad source trained model ($f_{\theta}, \psi_\phi$);\\ \quad  \quad  \quad
	  hyper-parameters $\lambda_{dis}$, $\lambda_{coh}$, $\alpha_0, \gamma$, $\epsilon$;\\ \quad  \quad  \quad
	  initialize: $\theta^1\!=\!\theta$, \{$\tilde{h}^{0}(x)=0,\, \forall x\in Q$\} 
    \State \textbf{Output}: Fine-tuned feature encoder parameter $\theta$

    \For{{i = 1} {\bf to} maxiters} 
    
	  \State $ p_{n} = \psi_{\phi} (concat (f_{\theta}(x^n_1),..,f_{\theta}(x^n_K))),\; $\\ \quad \quad \quad  $
	  \mbox{for}\ S_n=\{x^n_1,\cdots,x^n_K\},\forall n\!\in\!\{1,\cdots, N\}$
      \State $\nabla_\theta \mathcal{L}_{ft}  \leftarrow 0$
      \State Compute $\alpha_i$ using Eq.(\ref{eq:alpha})	
        \For{$x \in Q $}
     \State Compute  $\tilde{P^i}(Y|x)$ using Eq.(\ref{eq:hx})(\ref{eq:tildehx})(\ref{eq:pseudo}).
        \State$\nabla_\theta\mathcal{L}_{ft}  \leftarrow \nabla_\theta\mathcal{L}_{ft} 
	  +  \nabla_{\theta} \mathcal{L}^{tr}_{CE}  ((x,\tilde{P^i}(Y|x));\theta,\phi)$      
        \EndFor
   
        \State$ \nabla_\theta\mathcal{L}_{ft}  \leftarrow \nabla_\theta\mathcal{L}_{ft} 
	  + \nabla_{\theta } \mathcal{L}_{CE}  (S)   
	  + \lambda_{dis}\nabla_{\theta }\mathcal{L}_{dis}$ \\
   \quad \quad \quad \quad \quad
	  $+\lambda_{coh}\nabla_{\theta }\mathcal{L}^{ft}_{coh}$ 

	  \State $\theta^{i+1}  \leftarrow \theta^{i}  - \eta \nabla_{\theta=\theta^i}\mathcal{L}_{ft}$;\quad $\theta=\theta^{i+1}$ 
  \EndFor

  \end{algorithmic}
  \label{alg:target}
\end{algorithm}
  
%%%%%%%%%%%%%%%%%%%%%%%%%%
\begin{table*}[t]
\caption{Mean classification accuracy (95\% confidence interval in brackets) 
	for cross-domain 5-way 5-shot classification.
	$^\ast$ and $^\dagger$ denote the results reported in \cite{guo2020broader} and \cite{advTaskAug} respectively. Transductive methods are indicated using (T). Methods sharing query data via Batch Normalization are indicated using (BN). }
\setlength{\tabcolsep}{1pt}	
\resizebox{\textwidth}{!}{
\begin{tabular}{l|c|c|c|c|c|c|c|c}
\hline	
 & ChestX                    & CropDisea.                & ISIC                       & EuroSAT                  & Places                   & Planate                    & Cars                     & CUB                      \\
   \hline 
	MatchingNet$^\ast$\cite{vinyals2016matching} & $22.40_{(0.70)}$  & $66.39_{(0.78)}$ & $36.74_{(0.53)}$  & $64.45_{(0.63)}$     & \multicolumn{1}{|c|}{$-$}          &      \multicolumn{1}{|c|}{$-$}                         &     \multicolumn{1}{|c|}{$-$}                        &    \multicolumn{1}{|c}{$-$}                         \\
MAML (BN)$^\ast$\cite{finn2017model} & $23.48_{(0.96)}$    & $78.05_{(0.68)}$    & $40.13_{(0.58)}$   & $71.70_{(0.72)}$   & \multicolumn{1}{|c|}{$-$}          &      \multicolumn{1}{|c|}{$-$}                         &     \multicolumn{1}{|c|}{$-$}                        &    \multicolumn{1}{|c}{$-$}                         \\
   ProtoNet$^\ast$\cite{snell2017prototypical}     & $24.05_{(1.01)}$  & $79.72_{(0.67)}$  & $39.57_{(0.57)}$        & $73.29_{(0.71)}$        & $58.54_{(0.68)}$        &     $46.80_{(0.65)}$                      &     $41.74_{(0.72)}$                       &  $55.51_{(0.68)}$                    \\
   MetaOpt$^\ast$\cite{lee2019meta} & $22.53_{(0.91)}$         & $68.41_{(0.73)}$   & $36.28_{(0.50)}$   & $64.44_{(0.73)}$& \multicolumn{1}{|c|}{$-$}          &      \multicolumn{1}{|c|}{$-$}                         &     \multicolumn{1}{|c|}{$-$}                        &    \multicolumn{1}{|c}{$-$}                         \\

 RelNet (BN)$^\dagger$\cite{sung2018learning}     & $24.07_{(0.20)}$          & $72.86_{(0.40)}$     & $38.60_{(0.30)}$ & $65.56_{(0.40)}$   & $64.25_{(0.40)}$          & $42.71_{(0.30)}$            & $40.46_{(0.40)}$     & $56.77_{(0.40)}$     \\

   GNN$^\dagger$\cite{garcia2018fewshot}      & $23.87_{(0.20)}$       & $83.12_{(0.40)}$  & $42.54_{(0.40)}$   & $78.69_{(0.40)}$        & $70.91_{(0.50)}$  & $48.51_{(0.40)}$ & $43.70_{(0.40)}$   & $62.87_{(0.50)}$ \\
   %\hline 
   TPN (T)$^\dagger$\cite{liu2018learning}  & $22.17_{(0.20)}$ & $81.91_{(0.50)}$ & $45.66_{(0.30)}$ & $77.22_{(0.40)}$ & $71.39_{(0.40)}$  & $50.96_{(0.40)}$  & $44.54_{(0.40)}$ & $63.52_{(0.40)}$ \\
   \hline
MatchingNet+FWT$^\ast$\cite{Tseng2020CrossDomain}  & $21.26_{(0.31)}$         & $62.74_{(0.90)}$           & $30.40_{(0.48)}$     & $56.04_{(0.65)}$         & \multicolumn{1}{|c|}{$-$}          &      \multicolumn{1}{|c|}{$-$}                         &     \multicolumn{1}{|c|}{$-$}                        &    \multicolumn{1}{|c}{$-$}                         \\
  ProtoNet+FWT$^\ast$\cite{Tseng2020CrossDomain}  & $23.77_{(0.42)}$  & $72.72_{(0.70)}$  & $38.87_{(0.52)}$  & $67.34_{(0.76)}$         & \multicolumn{1}{|c|}{$-$}          &      \multicolumn{1}{|c|}{$-$}                         &     \multicolumn{1}{|c|}{$-$}                        &    \multicolumn{1}{|c}{$-$}                         \\
 RelNet+FWT (BN)$^\dagger$\cite{Tseng2020CrossDomain} & $23.95_{(0.20)}$          & $75.78_{(0.40)}$   & $38.68_{(0.30)}$           & $69.13_{(0.40)}$     & $65.55_{(0.40)}$         & $44.29_{(0.30)}$   & $40.18_{(0.40)}$         & $59.77_{(0.40)}$ \\
   GNN+FWT$^\dagger$\cite{Tseng2020CrossDomain}& $24.28_{(0.20)}$ & $87.07_{(0.40)}$  & $40.87_{(0.40)}$  & $78.02_{(0.40)}$  & $70.70_{(0.50)}$ & $49.66_{(0.40)}$ & $46.19_{(0.40)}$  & $64.97_{(0.50)}$       \\
  TPN+FWT (T) $^\dagger$\cite{Tseng2020CrossDomain} & $21.22_{(0.10)}$ & $70.06_{(0.70)}$  & $36.96_{(0.40)}$  & $65.69_{(0.50)}$ & $66.75_{(0.50)}$  & $43.20_{(0.50)}$ & $34.03_{(0.40)}$  & $58.18_{(0.50)}$\\
     %\hline 
     ATA $^\dagger$\cite{advTaskAug}  & $24.43_{(0.20)}$  & {$90.59_{(0.30)}$}   & {$45.83_{(0.30)}$}   & {$\mathbf{83.75}_{(0.40)}$} & {${75.48}_{(0.40)}$} & {$55.08_{(0.40)}$}   & {$49.14_{(0.40)}$} & {${66.22}_{(0.50)}$} \\
   %\hline 
     LRP-CAN  (T)  \cite{sun2021explanation}  &  \multicolumn{1}{|c}{$-$} & \multicolumn{1}{|c}{$-$}  & \multicolumn{1}{|c}{$-$}  & \multicolumn{1}{|c|}{$-$}       & 
       $\mathbf{76.90}_{(0.39)}$          &       $51.63_{(0.41)}$                        &     $42.57_{(0.42)} $                       &    $66.57_{(0.43)}$                        \\
    LRP-GNN \cite{sun2021explanation}  & \multicolumn{1}{|c}{$-$} & \multicolumn{1}{|c}{$-$}  & \multicolumn{1}{|c}{$-$}  & \multicolumn{1}{|c|}{$-$}       & 
       $74.45_{(0.47)}$          &       $54.46_{(0.46)}$                        &     $46.20_{(0.46)}$                       &    $64.44_{(0.48)}$                        \\
       CHEF\cite{adler2020cross}      & $24.72_{(0.14)}$& $86.87_{(0.20)}$ & $41.26_{(0.34)}$ & $74.15_{(0.27)}$    & \multicolumn{1}{|c|}{$-$}          &      \multicolumn{1}{|c|}{$-$}                         &     \multicolumn{1}{|c|}{$-$}                        &    \multicolumn{1}{|c}{$-$}                         \\
HVM\cite{du2022hierarchical}  & $\mathbf{27.15}_{(0.45)}$ & $87.65_{(0.35)}$  & $42.05_{(0.34)}$           & $74.88_{(0.45)}$       & \multicolumn{1}{|c|}{$-$}          &      \multicolumn{1}{|c|}{$-$}                         &     \multicolumn{1}{|c|}{$-$}                        &    \multicolumn{1}{|c}{$-$}                         \\
  APPL (T)  & $24.87_{(0.41)}$ & $\mathbf{92.51}_{(0.84)}$ & $\mathbf{46.28}_{(0.64)}$ & $79.78_{(0.78)}$ & $68.84_{(0.80)}$ & {$\mathbf{55.20}_{(0.58)}$} & $\mathbf{52.67}_{(0.42)}$ & $\mathbf{67.46}_{(0.78)}$\\  
\hline
\end{tabular}} 
\label{table:5shot}
\end{table*}
%%%%%%%%%

%%%%%%%
With the soft pseudo-labels predicted by 
the current prototype-based model ($\theta^i,\phi$),
the query instances can be deployed through a cross-entropy loss to 
guide the further update, i.e., fine-tuning, of the feature encoder $f_{\theta}$. 
However, using all pseudo-labeled query instances 
may lead to noisy updates and negatively impact the model
due to the low-confidence predictions of the pseudo-labels.  
Therefore, we choose to only employ query instances with high prediction confidence scores
that are larger than a predefined threshold $\epsilon$ 
and compute the query-based cross-entropy loss as follows:
\begin{equation}\label{eq:ce-psudo}
\mathcal{L}^{tr}_{CE}(Q)= 
\sum_{x \in Q}
     \begin{cases}
	     \mathcal{L}_{CE}(x,\tilde{P}^{i}(Y|x);\theta,\phi) &\text{if }\max(\tilde{P}^{i}(Y|x))>\epsilon\\
       0 &\text{otherwise}
     \end{cases}
\end{equation}
where $\tilde{P}^{i}(Y|x)$ denotes the soft pseudo-label vector computed via Eq.(\ref{eq:pseudo}) 
for query instance $x$, 
while the maximum predicted probability, $\max(\tilde{P}^{i}(Y|x))$, is used as the prediction confidence score for 
query instance $x$.
Here $\mathcal{L}_{CE}(x,\tilde{P}^{i}(Y|x);\theta,\phi)$ denotes the cross-entropy loss
computed over the soft pseudo-labeled pair $(x,\tilde{P}^{i}(Y|x))$
with $f_\theta$ and $\psi_\phi$. 

In addition to the cross-entropy loss on both the support 
and query instances, we also use the prototype regularization losses, 
$\mathcal{L}_{dis}$ and $\mathcal{L}_{coh}$,
introduced in the meta-training phase to guide the fine-tuning process. 
Since the true labels of the query instances are unknown in the meta-testing phase,
we modify the prototype cohesive loss $\mathcal{L}_{coh}$ and compute it on the support instances instead: 
\begin{equation}
            \mathcal{L}^{ft}_{coh} = \sum\nolimits_{n=1}^{N}\sum\nolimits_{x \in S_{n}} d(p_n,f_\theta(x) ),
\end{equation}
where $S_{n}$ is the set of support instances from class n. 
Overall, the feature encoder is fine-tuned by minimizing the following joint loss function
with gradient descent:
\begin{equation}\label{eq:finetune}
     	\min_{\theta} \;  \mathcal{L}_{ft} = \mathcal{L}_{CE}(S)+ \mathcal{L}^{tr}_{CE}(Q) + \lambda_{dis} \mathcal{L}_{dis} 
      +\lambda_{coh}       \mathcal{L}^{ft}_{coh}%.
 \end{equation}
This fine-tuning procedure is summarized in 
Algorithm \ref{alg:target}. 

%%%%%%%%%%%%%%%%%%%%%%%%%%%%%%%%%%%%%%%%%%%%%%%%%%

\section{Experiments}

In this section, we present our experimental settings and results, 
comparing the proposed APPL method with a great set of methods under the CDFSL setting. 

\subsection{Experimental Setup}

\noindent\textbf{Datasets\ \ }
We conducted comprehensive experiments on eight cross-domain few-shot learning (CDFSL) benchmark datasets.
We use MiniImageNet \cite{vinyals2016matching} as the single source domain dataset,
and use the following eight datasets as the target domain datasets: 
CropDiseases \cite{mohanty2016using}, EuroSAT \cite{helber2019eurosat}, 
ISIC \cite{tschandl2018ham10000}, 
ChestX \cite{wang2017chestx}, 
CUB \cite{wah2011caltech}, Cars \cite{krause20133d}, 
Places \cite{zhou2017places} and Planate \cite{van2018inaturalist}.
We use the same train/val/test split as \cite{guo2020broader}.  
We select the hyperparameters based on the test accuracy on the MiniImageNet validation set.\\

\noindent\textbf{Implementation Details}
We use ResNet10 \cite{he2016deep} as our backbone network and use a simple network made up of a single linear layer followed by ReLU activation to represent PCN. 
We train our prototype-based prediction model (feature encoder and PCN)
on the source domain for 400 epochs
with 100 meta-training tasks and 15 query instances per class. 
Adam optimizer with weight decay of 1e-2 and learning rate of 1e-6 is used to train the APPL.
The trade-off parameters $\lambda_{dis}$ and $\lambda_{coh}$ are set to 0.1 and 1e-3 respectively. 
The proposed APPL is evaluated on 600 randomly selected few-shot learning tasks in each target domain.
We fine-tune the feature encoder for 100 iterations 
for each task with a learning rate of 1e-2.
For the fine-tuning hyperparameters, 
$\alpha_0$, $\gamma$ and $\epsilon$ take the values of
$0.5$, $0.99$ and $0.4$ respectively. 

%%%%%%%%%%%%%%%%%%%%%%%%%%%%%%%%%%%%%
\subsection{Comparison Results}
\begin{table*}[t]
\caption{Mean classification accuracy (95\% confidence interval within brackets) 
	for cross-domain 5-way 20-shot and 50-shot classification. 
	$^\ast$ denotes results reported in \cite{guo2020broader}. Transductive methods are indicated using (T). Methods sharing query data via Batch Normalization are indicated using (BN).}
\setlength{\tabcolsep}{2pt}	
\resizebox{\textwidth}{!}{
\begin{tabular}{l|c|c|c|c|c|c|c|c  }
\hline	
&   \multicolumn{2}{c|}{ChestX}         &   \multicolumn{2}{c|}{CropDiseases}   &           \multicolumn{2}{c|}{ISIC}             &     \multicolumn{2}{c}{EuroSAT}                  \\
  &  20-shot & 50-shot &  20-shot & 50-shot &  20-shot & 50-shot &  20-shot & 50-shot \\   
   
   \hline 
 MatchingNet$^\ast$\cite{vinyals2016matching} & $23.61_{(0.86)}$  &    $22.12_{(0.88)}$   & $76.38_{(0.67)}$ &    $58.53_{(0.73)}$ & $45.72_{(0.53)}$ &    $54.58_{(0.65)}$ & $77.10_{(0.57)}$ &    $54.44_{(0.67)}$          \\
   MAML (BN)$^\ast$\cite{finn2017model} & $27.53_{(0.43)}$    &    $-$   & $89.75_{(0.42)}$    &    $-$   & $52.36_{(0.57)}$    &   $-$   & $81.95_{(0.55)}$   &    $-$    \\
 ProtoNet$^\ast$\cite{snell2017prototypical}  & $28.21_{(1.15)}$   &    $29.32_{(1.12)}$   & $88.15_{(0.51)}$   &    $90.81_{(0.43)}$   & $49.50_{(0.55)}$        &    $51.99_{(0.52)}$   & $82.27_{(0.57)}$  &    $80.48_{(0.57)}$       \\
   MetaOpt$^\ast$\cite{lee2019meta}  & $25.53_{(1.02)}$   &    $29.35_{(0.99)}$   & $82.89_{(0.54)}$    &    $91.76_{(0.38)}$   & $49.42_{(0.60)}$   &    $54.80_{(0.54)}$   & $79.19_{(0.62)}$   &    $83.62_{(0.58)}$            \\

RelNet (BN)$^\ast$\cite{sung2018learning}     & $26.63_{(0.92)}$     &    $28.45_{(1.20)}$   & $80.45_{(0.64)}$  &    $85.08_{(0.53)}$   & $41.77_{(0.49)}$  &    $49.32_{(0.51)}$   & $74.43_{(0.66)}$  &    $74.91_{(0.58)}$     \\
\hline
 MatchingNet+FWT$^\ast$\cite{Tseng2020CrossDomain}  & $23.23_{(0.37)}$        &    $23.01_{(0.34)}$   & $74.90_{(0.71)}$         &    $75.68_{(0.78)}$   & $32.01_{(0.48)}$     &    $33.17_{(0.43)}$   & $63.38_{(0.69)}$  &    $62.75_{(0.76)}$     \\
 ProtoNet+FWT$^\ast$\cite{Tseng2020CrossDomain}   & $26.87_{(0.43)}$   &    $30.12_{(0.46)}$   & $85.82_{(0.51)}$   &    $87.17_{(0.50)}$   & $43.78_{(0.47)}$   &    $49.84_{(0.51)}$   & $75.74_{(0.70)}$   &    $78.64_{(0.57)}$        \\
 RelNet+FWT(BN)$^\ast$\cite{Tseng2020CrossDomain} & $26.75_{(0.41)}$    &    $27.56_{(0.40)}$   & $78.43_{(0.59)}$    &    $81.14_{(0.56)}$   & $43.31_{(0.51)}$  &    $46.38_{(0.53)}$   & $69.40_{(0.64)}$   &    $73.84_{(0.60)}$  \\
 CHEF\cite{adler2020cross}      & $29.71_{(0.27)}$  &    $31.25_{(0.20)}$   & $94.78_{(0.12)}$  &    $96.77_{(0.08)}$   & $54.30_{(0.34)}$  &    $60.86_{(0.18)}$   & $83.31_{(0.14)}$ &    $86.55_{(0.15)}$             \\
    HVM\cite{du2022hierarchical}  & ${30.54}_{(0.47)}$  &    $32.76_{(0.46)}$   & $95.13_{(0.35)}$   &    $97.83_{(0.33)}$   & $54.97_{(0.35)}$            &    $61.71_{(0.32)}$   & $84.81_{(0.34)}$      &    $87.16_{(0.35)}$    \\
  APPL (T)  & $\mathbf{30.75}_{(0.41)}$  &    $\mathbf{33.14}_{(0.88)}$   & $\mathbf{95.77}_{(0.53)}$  &    $\mathbf{98.14}_{(0.56)}$   & $\mathbf{57.97}_{(0.73)}$  &    $\mathbf{62.17}_{(0.43)}$   & $\mathbf{88.60}_{(0.85)}$  &    $\mathbf{89.75}_{(0.76)}$   \\  
\hline
\end{tabular}} 
\label{table:20shot}
\end{table*}
\begin{table*}[t]
\caption{
	Ablation study results 
	for cross-domain 5-way 5-shot classification tasks.}
	\setlength{\tabcolsep}{2pt}	
\resizebox{\textwidth}{!}{
\begin{tabular}{l|c|c|c|c|c|c|c|l}
\hline	
   & ChestX                    & CropDisea.                & ISIC                       & EuroSAT                  & Places                   & Planate                    & Cars                     & CUB                      \\
   \hline 
  ProtoNet                   & $24.05_{(1.01)}$  & $79.72_{(0.67)}$  & $39.57_{(0.57)}$        & $73.29_{(0.71)}$        & $58.54_{(0.68)}$        &         $46.80_{(0.65)}$                       &     $41.74_{(0.72)}$                       &  $55.51_{(0.68)}$                    \\
\hline
APPL  & $\mathbf{24.87}_{(0.41)}$ & $\mathbf{92.51}_{(0.84)}$ & $\mathbf{46.28}_{(0.64)}$ & $\mathbf{79.78}_{(0.78)}$ & $\mathbf{68.84}_{(0.80)}$ & {${55.20}_{(0.58)}$} & $\mathbf{51.67}_{(0.42)}$ & $\mathbf{67.46}_{(0.78)}$\\ 
$\; - \text{w/o } \psi_{\phi}$   &   $22.33_{(0.56)}$&	$89.11_{(0.66)}$  & $43.99_{(0.68)}$ &	$77.99_{(0.68)}$ &$67.57_{(0.33)}$&	$53.69_{(0.60)}$&	$50.30_{(0.81)}$&	$64.03_{(0.97)}$\\

$\; - \text{w/o } \mathcal{L}_{dis}$   & ${24.84}_{(0.69)}$ & $91.31_{(0.72)}$ &	$43.23_{(0.82)}$&	$78.25_{(0.76)}$ &	$66.18_{(0.49)}$&	$54.56_{(0.84)}$&	$51.56_{(0.85)}$&	$60.45_{(0.85)}$\\  

$\; - \text{w/o } \mathcal{L}_{coh}$   & $23.77_{(0.68)}$&	$90.19_{(0.76)}$&	$43.69_{(0.49)}$&	$79.53_{(0.71)}$ &	$65.89_{(0.47)}$&	$51.62_{(0.77})$&	$50.32_{(0.81)}$&	$60.96_{(0.85)}$\\  

$\; - \text{w/o } \mathcal{L}_{CE}(S)$  & $21.08_{(0.42)}$ &	$59.15_{(0.59)}$&	$26.90_{(0.78)}$ & $49.95_{(0.80)}$ &	$34.79_{(0.85)}$ &	$25.72_{( 0.77)}$	& $27.19_{(0.87)}$&	$30.12_{(0.74)}$\\

$\; - \text{w/o } \mathcal{L}^{tr}_{CE}(Q)$  & $22.36_{(0.42)}$ &	$88.24_{(0.86)}$&	$42.14_{(0.59)}$&	$77.63_{(0.65)}$&	$67.14_{(0.82)}$& 	$\mathbf{55.69}_{(0.60)}$&	$51.59_{(0.82)}$&	$60.22_{(0.85)}$\\ 

 $\; - \text{w/o}$ WMA  & $21.15_{(0.45)}$ &	$90.12_{(0.87)}$&	$44.95_{(0.62)}$&	$77.12_{(0.67)}$&	$67.68_{(0.83)}$& 	${54.76}_{(0.62)}$&	$49.94_{(0.85)}$&	$65.87_{(0.87)}$\\ 
\hline
\end{tabular}} 
\label{table:ablation}
\end{table*}
%%%%%%%%%%%%%%%%%
\subsubsection{Learning with Few Shots}
We first evaluate the performance of the proposed APPL method on the common 
cross-domain 5-way 5-shot classification tasks.
We compare APPL with both a set of representative FSL methods 
(MatchingNet \cite{vinyals2016matching}, MAML \cite{finn2017model}, ProtoNet \cite{snell2017prototypical}, RelationNet \cite{sung2018learning}, MetaOpt \cite{lee2019meta}, GNN \cite{garcia2018fewshot} and TPN \cite{liu2018learning}) and five state-of-the-art CDFSL methods 
(FWT \cite{Tseng2020CrossDomain}, ATA \cite{advTaskAug}, LRP \cite{sun2021explanation}, CHEF \cite{adler2020cross} and HVM \cite{du2022hierarchical}). 
FWT has been applied jointly with five standard FSL methods: MatchingNet, ProtoNet, RelationNet, GNN and TPN. LRP has been applied jointly with two standard FSL methods: GNN and Cross-attention network (CAN). 
The comparison results are presented in Table~\ref{table:5shot}, 
where the top part of the table reports the results of the standard FSL methods and 
the bottom part reports the results of the CDFSL methods. 

We can see that the CDFSL methods 
(ATA, CHEF, HVM, LRP and APPL) 
designed specifically to handle vast differences between source and target domains 
perform largely better than the standard FSL works for in-domain settings. 
FWT however only produces improvements in most cases over its base RelationNet. 
Notably, the proposed APPL outperforms all standard FSL methods 
including ProtoNet and ProtoNet+FWT on all the eight datasets. 
In particular, its performance gain over ProtoNet is remarkable, 
exceeding 10\% on four out of the eight datasets,
which highlights the importance of the prototype learning network. 
In addition, APPL outperforms all the CDFSL 
methods on five datasets, 
and produces the second best results on the two datasets. 
These results demonstrate the effectiveness of the proposed APPL method 
for cross-domain few-shot learning. 

%%%%%%%%%%%%%%%%%

\subsubsection{Learning with Higher Shots}
We further investigated CDFSL with higher-shot tasks in the target domain. 
In particular, we evaluate APPL with cross-domain 5-way 20-shot and 5-way 50-shot learning tasks 
on 4 target-domain datasets (ChestX, CropDiseases, ISIC and EuroSAT).
To handle higher-shot problems and increase the scalability of APPL, 
we extend APPL by adding a clustering function $g$  prior to the PCN component. By adopting this clustering approach, we ensure that the number of learnable parameters of our proposed PCN is independent of the number of shots which highlights the scalability of our proposed approach.
The $g$ function clusters the support instances in each class into $K'=5$ clusters
based on their learned embeddings. The obtained cluster centroid vectors are then concatenated as input for PCN. 
We compared APPL with both standard FSL methods and several CDFSL methods, 
and the results are presented in Table~\ref{table:20shot}, where the top part of the table presents the results of the FSL methods and the bottom part presents the results of the CDFSL methods.

As observed from the results, the methods designed for CDFSL (CHEF, HVM, and APPL) continue to demonstrate superior performance compared to the FSL methods. The performance gains of APPL over Protonet and Protonet+FWT are remarkable exceeding 6\% and 9\% on three datasets (CropDiseases, ISIC and EuroSAT) in the cases of 20-shot and 50-shot respectively.
Moreover, the APPL method consistently achieved the best performance across all four datasets for both the traditional FSL methods and the other CDFSL methods for the 20-shot and 50-shot cases.
These results again validate the effectiveness of APPL for cross-domain few-shot learning
and demonstrate its capacity in handling cross-domain higher-shot learning problems. Two factors account for our proposed method's good performance in the case of higher shots: First, benefiting from the proposed WMA self-training approach in the target domain, we are able to generate more accurate pseudo-labels with higher shots, which enables our model to obtain better results. Second, we conduct clustering over the embeddings of support instances to generate class centroids with higher shots, which can eliminate some noisy information and allow the PCN to learn the most representative features.

%%%%%%%%%%%%%%%%%%%%%%%%%%%%%%

\begin{figure*}
\centering
\begin{subfigure}{0.24\textwidth}
\centering
\includegraphics[width = \textwidth]{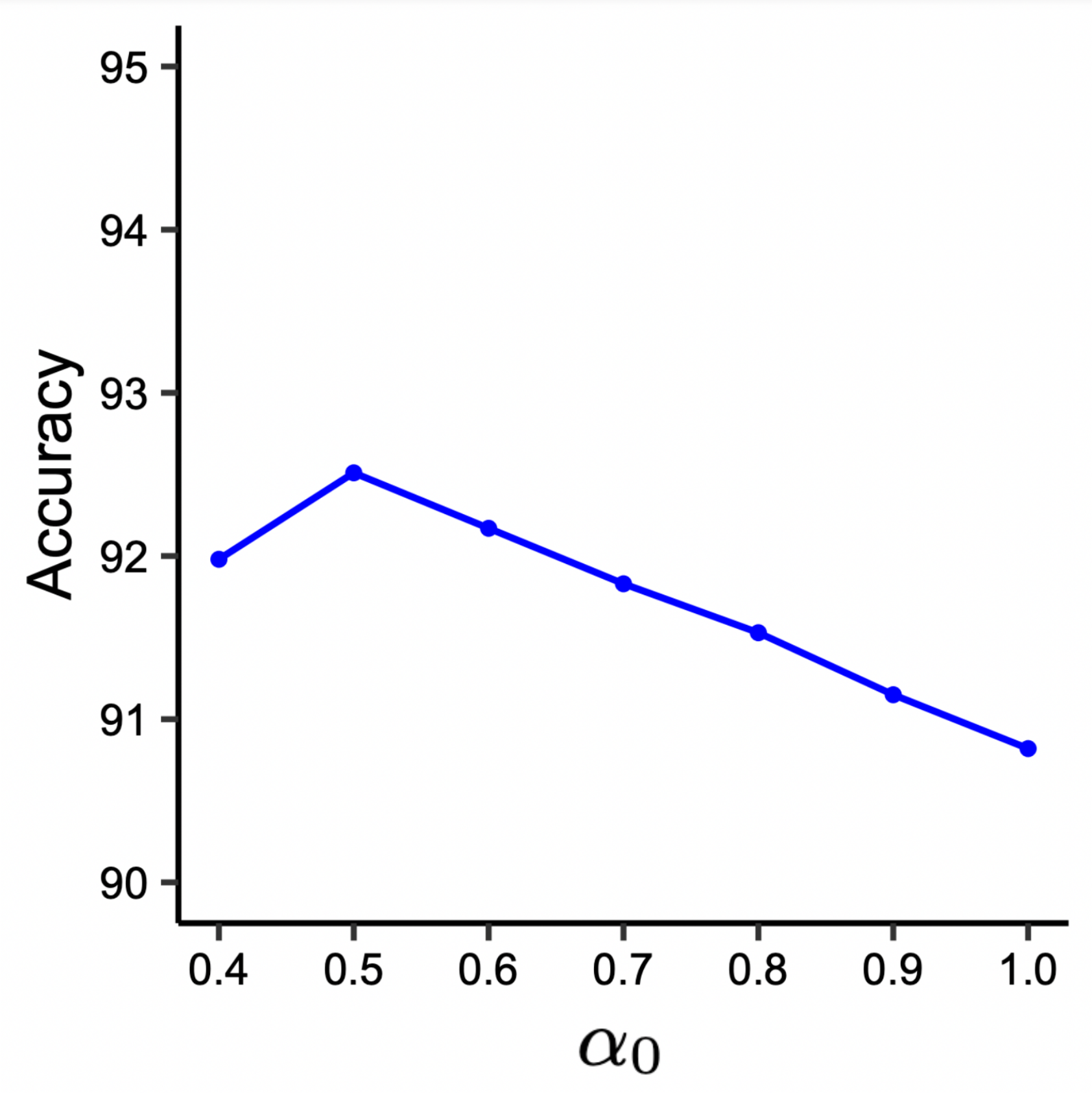}
\caption{$\alpha_0$}
\end{subfigure}
\begin{subfigure}{0.24\textwidth}
\centering
\includegraphics[width = \textwidth]{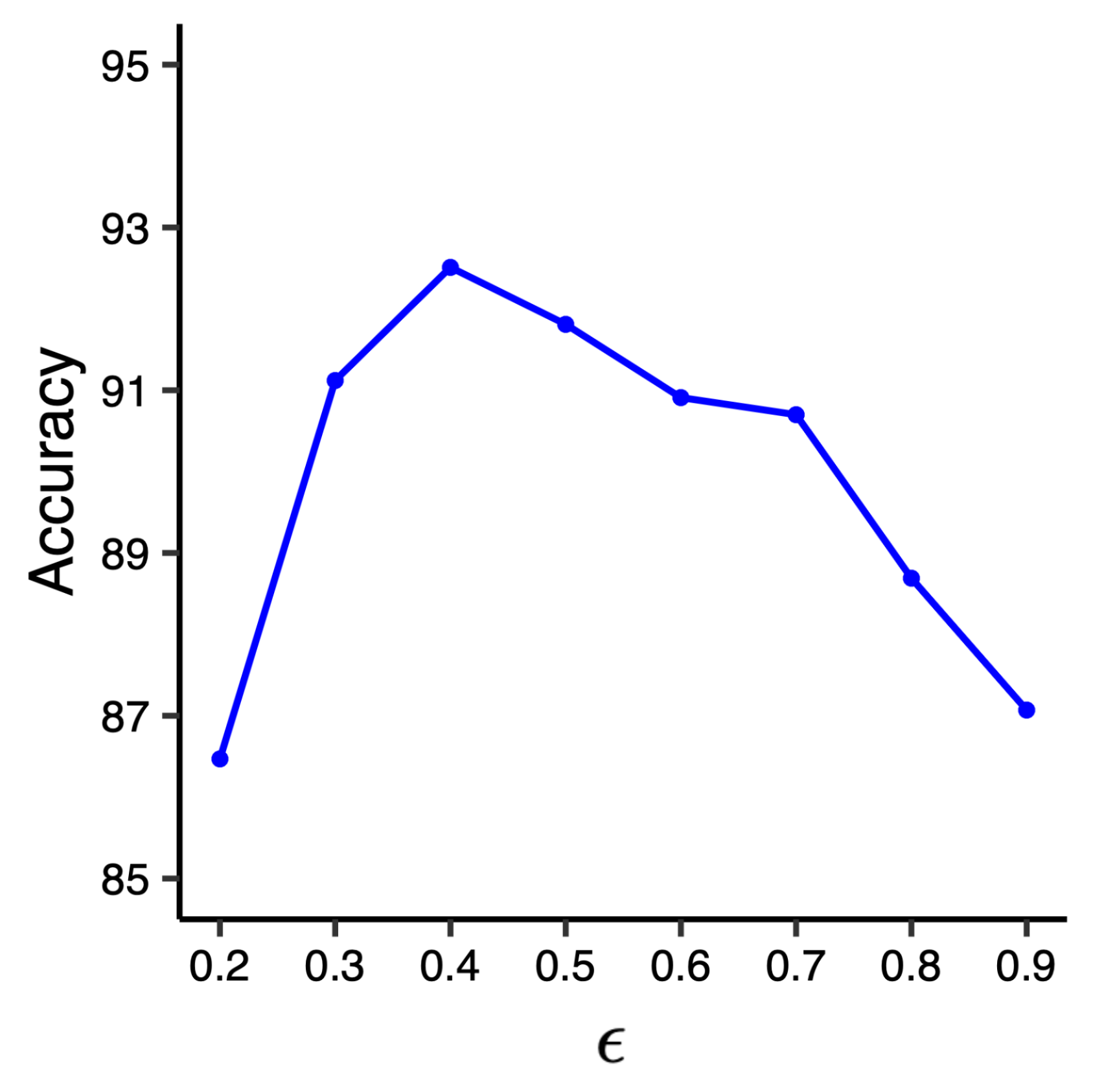} 
\caption{$\epsilon$}
\end{subfigure}
\begin{subfigure}{0.24\textwidth}
\centering
\includegraphics[width = \textwidth]{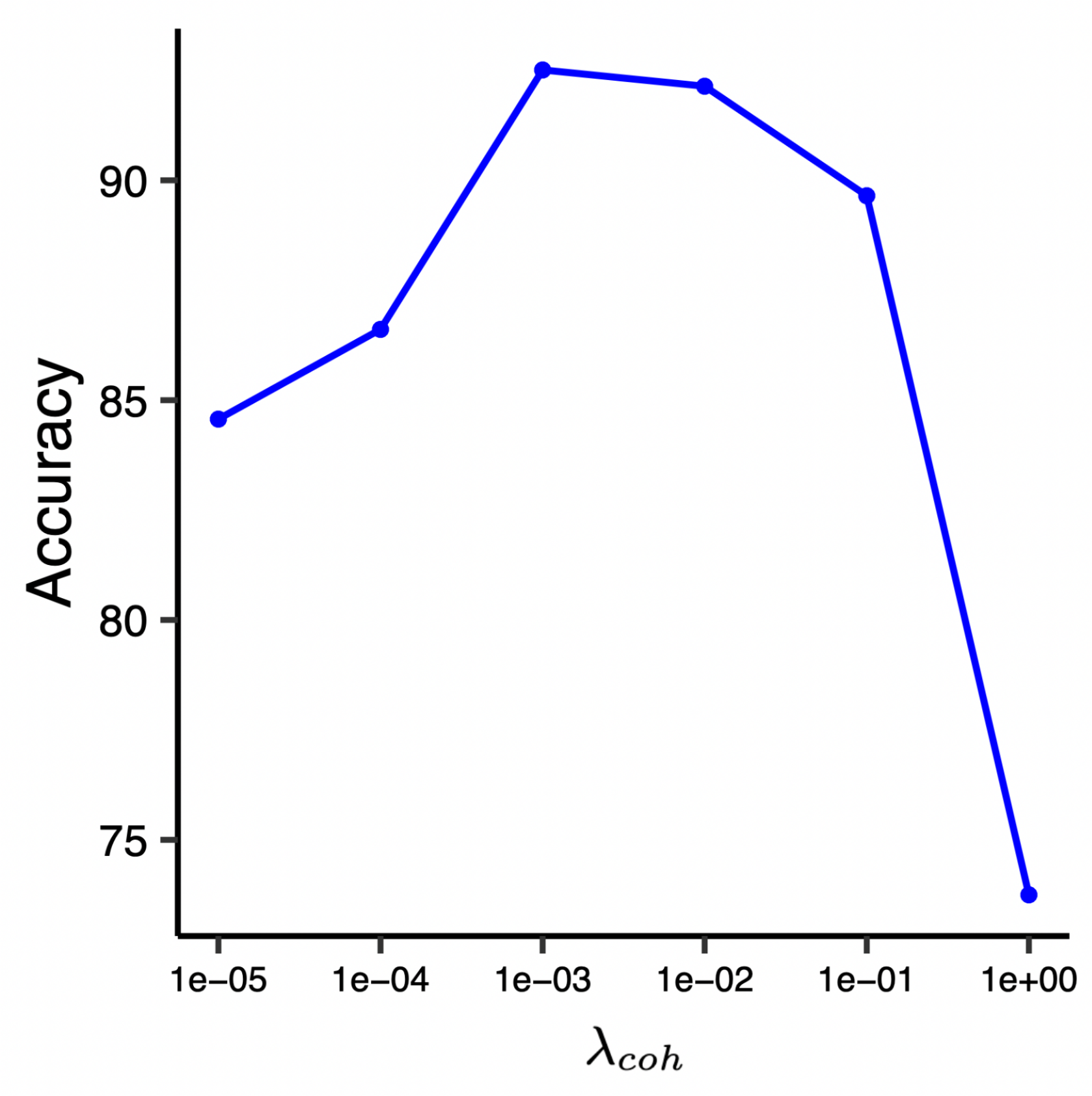}
\caption{$\lambda_{coh}$}
\end{subfigure}
\begin{subfigure}{0.24\textwidth}
\centering
\includegraphics[width = \textwidth]{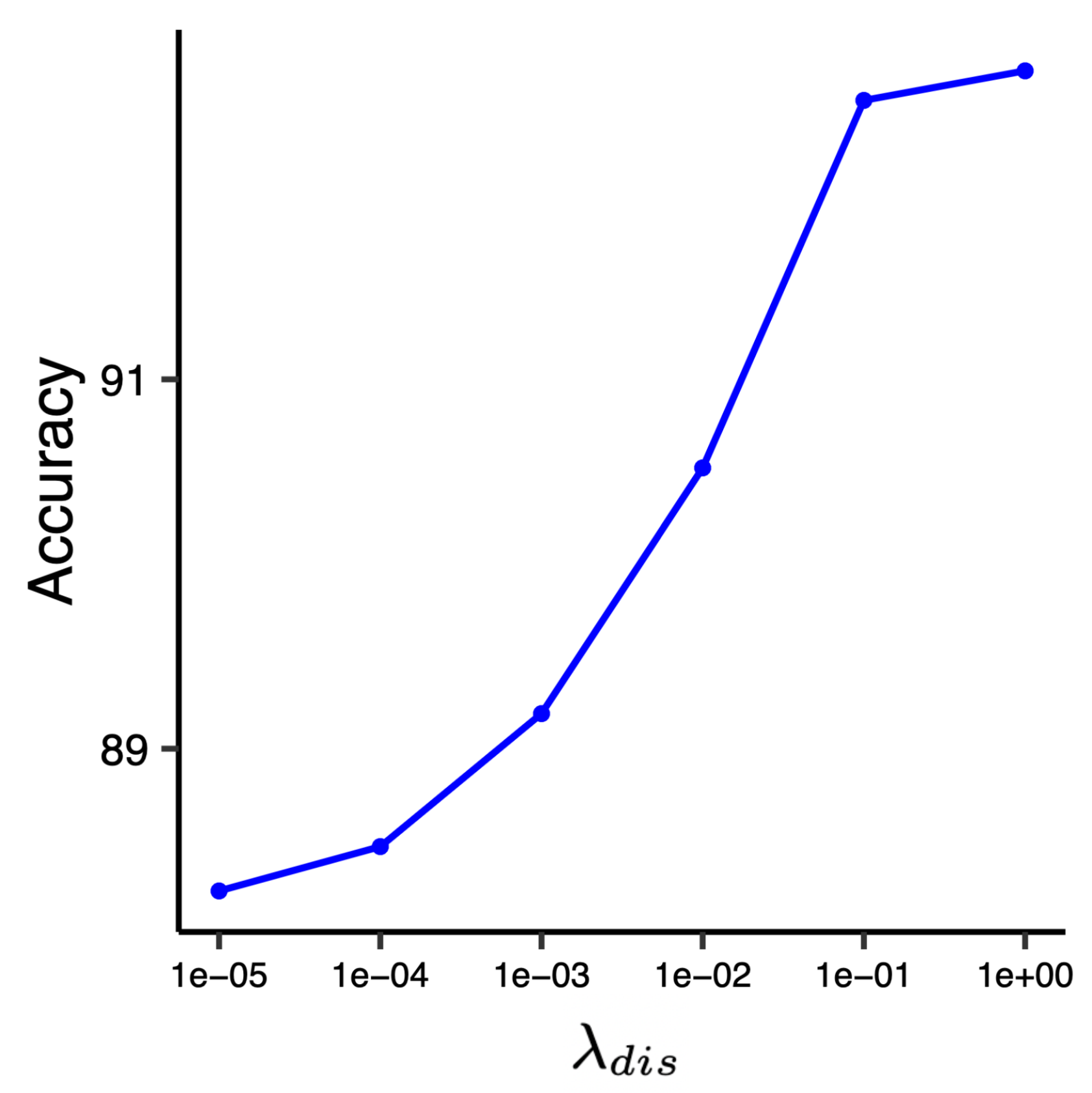}
\caption{$\lambda_{dis}$}
\end{subfigure}
\caption{Sensitivity analysis for the proposed method on hyper-parameters $\alpha_0$, $\epsilon$, $\lambda_{coh}$ and $\lambda_{dis}$ on CropDisease dataset under cross-domain 5-way 5-shot task: (a) $\alpha_0$, (b) $\epsilon$, (c)  $\lambda_{coh}$, (d) $\lambda_{dis}$. }
\label{fig:hyper_sen}
\end{figure*}

%%%%%%%%%%%%%%%%%%%%%%%%%%%%%%%%%%
\subsection{Ablation Study}

To investigate the importance of each component of our APPL approach, 
we conducted an ablation study to compare APPL with its six variants:
(1) ``$- \text{w/o } \psi_{\phi}$'', which drops PCN and replaces it with  
a simple average of the support instances of each class. 
(2) ``$- \text{w/o } \mathcal{L}_{dis}$'' and 
(3) ``$- \text{w/o } \mathcal{L}_{coh}$'', which drop 
 the $ \mathcal{L}_{dis}$ loss and $\mathcal{L}_{coh}$ loss respectively.
(4) ``$- \text{w/o } \mathcal{L}_{CE}(S)$'', 
which drops the cross-entropy loss over the support instances in fine-tuning. 
(5) ``$- \text{w/o } \mathcal{L}^{tr}_{CE}(Q)$'', 
which drops the cross-entropy loss over the query instances 
and hence the WMA self-training component in fine-tuning. 
(6) ``$- \text{w/o}$ WMA'', 
which drops the weighted moving average when determining the pseudo-labels, and instead, leverages the predictions generated by the model at the current iteration only to generate the pseudo-labels
(7) ``ProtoNet'', which can be considered as a variant of APPL that drops PCN, WMA self-training, $\mathcal{L}_{dis}$ and $\mathcal{L}_{coh}$.

We compared APPL with its six variants on the cross-domain 5-way 5-shot setting 
on all the eight datasets, 
and the results are reported in Table \ref{table:ablation}. 
We can see that
APPL outperforms all the other variants on almost all datasets.
The ``$- \text{w/o } \mathcal{L}_{CE}(S)$'' variant produced the largest performance drop among all variants, 
which highlights the importance of the few labeled support instances 
for fine-tuning in the target domain. 
The performance degradation for ProtoNet and ``$- \text{w/o } \psi_{\phi}$''
highlights the importance of the proposed PCN component.
In addition, ``$- \text{w/o } \psi_{\phi}$'' outperforms ProtoNet, 
which underlines the performance gain obtained by 
using pseudo-labeled query instances with WMA self-training and the prototype regularization losses 
in the absence of PCN. 
The other four variants, ``$- \text{w/o } \mathcal{L}_{dis}$'', ``$- \text{w/o } \mathcal{L}_{coh}$'', ``$- \text{w/o } \mathcal{L}^{tr}_{CE}(Q)$'', and ``$- \text{w/o}$ WMA'' also perform worse than APPL,
which verifies the contributions of 
the two prototype regularization loss terms and
the WMA self-training component respectively. 

%%%%%%%%%
\section{Hyper-parameter Sensitivity Analysis}
To demonstrate the effect of the hyper-parameters of APPL, we summarize the results of various studies in Figure \ref{fig:hyper_sen}.
The figure shows the performance of APPL on CropDisease dataset under the cross-domain 5-way 5-shot task as we modify each hyper-parameter separately.  
As seen from the results, it is clear that APPL is not sensitive to the choice of value for $\alpha_0$, as the performance of APPL is stable across all values of $\alpha_0$. 
As for $\lambda_{coh}$, too large or too small a value will pull the query/support instances to the prototypes more strongly or more weakly, thus having a negative impact on the results.
A reasonable value between 1e-1 and 5e-3 is required to obtain reasonable performance. 
In the case of $\lambda_{dis}$, the experimental results improve as the value of $\lambda_{dis}$ increases. And the results become stable when the parameter reaches 1e-1.
We believe that this is because $\lambda_{dis}$ controls the scale of pushing the prototypes of different classes away from each other so that when the distance between the prototypes reach a certain threshold, embedding is no longer susceptible to mutual influence.

Finally, it is worth noting that $\epsilon$ is an important hyperparameter. $\epsilon$ represents the degree of certainty of the pseudo-labels' predictions utilized in fine-tuning. When $\epsilon$ is too small, more unlabeled samples are used in fine-tuning with their noisy pseudo-labels, and when $\epsilon$ is too large, less unlabeled samples are used in fine-tuning with less noisy pseudo-labels.
Experimental results show that when the intermediate value of 0.4 is selected, the number and accuracy of pseudo-labels can be balanced and the best results for this model were obtained.

\section{Conclusion}

In this paper, we proposed 
a novel Adaptive Parametric Prototype Learning (APPL) method to address the cross-domain few-shot learning problem.
APPL meta-trains an adaptive prototype calculator network in the source domain 
to learn more discriminative and representative class prototypes 
from various support instance layouts,
which can then guide the feature encoder to adapt to the target domain through fine-tuning. 
Moreover, a weighted-moving-average self-training approach is adopted
to enhance fine-tuning
by exploiting the unlabeled query instances in the target domain 
to mitigate domain shift and avoid overfitting to support instances. 
Experimental results on eight benchmark cross-domain few-shot classification datasets
demonstrate that APPL achieves the state-of-the-art performance for CDFSL.

{\small
\bibliographystyle{ieee_fullname}
\bibliography{ref}
}

\end{document}